# Smart Traffic Management of Vehicles using Faster R-CNN based Deep Learning Method

Arindam Chaudhuri
Associate Professor Data Analytics
Great Lakes Institute of Management Chennai
arindamchaudhuri.b@greatlakes.edu.in

**Abstract**

With constant growth of civilization and modernization of cities all across the world since past few centuries smart traffic management of vehicles is one of the most sorted after problem by research community. It is a challenging problem in computer vision and artificial intelligence domain. Smart traffic management basically involves segmentation of vehicles, estimation of traffic density and tracking of vehicles. The vehicle segmentation from traffic videos helps realization of niche applications such as monitoring of speed and estimation of traffic. When occlusions, background with clutters and traffic with density variations are present, this problem becomes more intractable in nature. Keeping this motivation in this research work, we investigate Faster R-CNN based deep learning method towards segmentation of vehicles. This problem is addressed in four steps viz minimization with adaptive background model, Faster R-CNN based subnet operation, Faster R-CNN initial refinement and result optimization with extended topological active nets. The computational framework uses ideas of adaptive background modeling. It also addresses shadow and illumination related issues. Higher segmentation accuracy is achieved through topological active net deformable models. The topological and extended topological active nets help to achieve stated deformations. Mesh deformation is achieved with minimization of energy. The segmentation accuracy is improved with modified version of extended topological active net. The experimental results demonstrate superiority of this computational framework.

## 1 Introduction

We are facing the challenge of continuous increase in traffic density across different cities of world. The present world population is heavily dependent on vehicles from smooth commutation purposes. This results from constant urbanization of rural areas. In order to address this problem, smart traffic management is studied by researchers as traffic regulations solution.

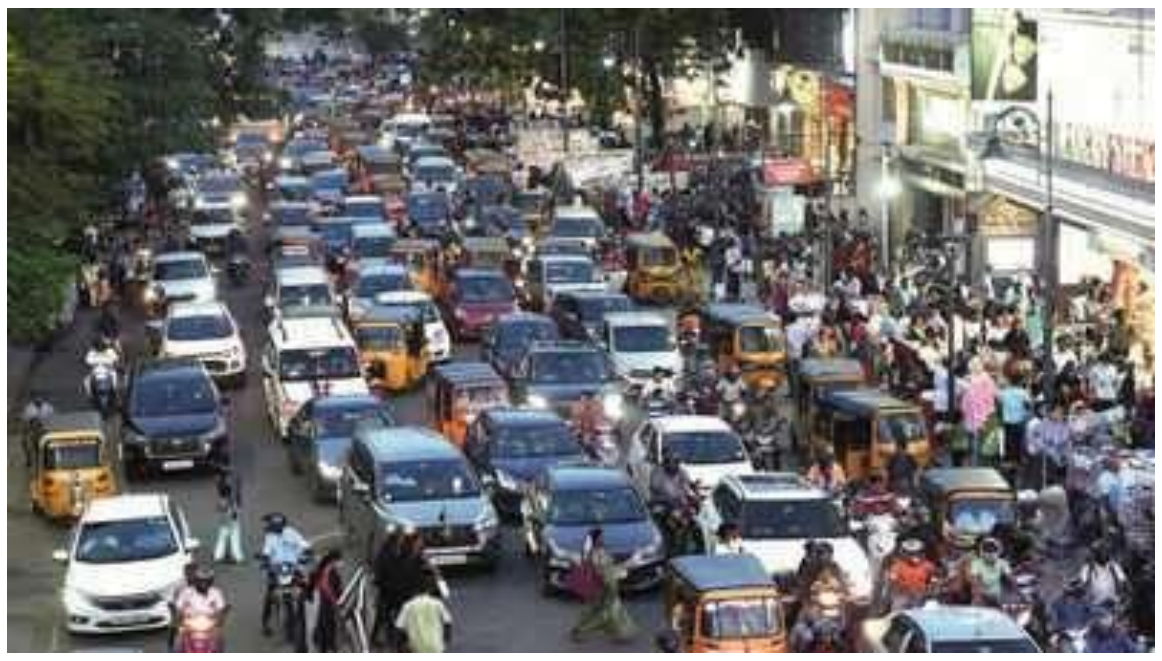

Figure 1: A real life crowded place from Indian traffic

Human vision system has the capability to perform complex tasks very reliably and accurately. Humans detect wide spectrum of objects very easily. With recent developments in computer vision coupled with huge data sets, better algorithms and faster GPUs, precision and accuracy of object detection and classification algorithms [1], [2] have increased to an appreciable amount. For traffic monitoring vehicle localization efficiency is very important. The vehicles from Indian traffic are presented in Figure 1. For public safety autonomous vehicle detection methods [3] are in place which detect traffic objects in order to achieve correct decisions.

Smart traffic management allows us to take care of various traffic related issues in an optimum manner. This is achieved using computer vision and image processing. Segmentation of vehicle acts as significant enabler in smart traffic management functionality. With occlusions, congestion as well as other environmental factors this problem becomes more severe.

For smooth and effective traffic regulations, smart traffic management is always considered as low cost solution. Congestions, emergency vehicles transport, accidents and traffic related violations are easily managed through latest Artificial Intelligence algorithms. Vehicle segmentation is an important area towards smart traffic management. Some other activities here include traffic estimation, speed control and vehicle tracking. During occlusions, fog, haze, clutters and heavy traffic situations things become more complicated.

With the growth of deep learning networks vehicle detection is being studied deeply considering traffic congestion and driving safety [4]. Vehicle localization is a crucial problem [5] in order to develop intelligent and autonomous systems. The detection of abnormalities arising from traffic violations leads to the problem of vehicle localization. This is a significant application catering the needs for variety of traffic related problems. The traffic surveillance has been a major concern in densely populated geographical areas. These days' surveillance systems are well equipped with traffic flow data where various traffic patterns are recorded. Some notable applications here include many smart cities based applications. Different versions of deep learning network based methods are in place which have provided considerable benefits for these applications [3].

In this research work, we investigate Faster R-CNN based deep learning method towards segmentation of vehicles. This problem is addressed in four steps [5]: (a) minimization with adaptive background model (b) Faster R-CNN based subnet operation (c) Faster R-CNN initial refinement and (d) result optimization with extended topological active nets. Adaptive gain function is used adaptive background modeling. The gain function compensates for shadow and illumination issues. Higher accuracy in segmentation is provided by topological active net deformable model in various situations. Deformable models provide various curvatures with respect to image surfaces. The smoothness from deformation is achieved through several forces considering objects of interest. The topological and extended topological active nets are used here in order to achieve stated deformations. This helps in fitting objects on 2D surface mesh of image. The deformation of mesh is achieved with minimization of energy. For problems with complex shapes energy is changed through extended topological active net with certain thresholds. The segmentation accuracy is increased with improved version of extended topological active net. This solution has combined effect from all models using which better segmentation boundaries are achieved.

The performance of stated method is compared with respect to some important metrics such as precision and recall. This method provides better segmentation performance in comparison to other methods. The experimental hypothesis is justified with benchmarked datasets. The model provides appreciable results for different sets of image datasets. This chapter is structured as follows. The literature review is presented in section 2. In section 3 highlights detailed discussion on computational method. The simulation results are shown in section 4. Finally, conclusion is provided in section 5.

## 2 Literature Review

During recent decades' smart traffic management involving vehicle detection has gained considerable attention. Some of the notable works are discussed here.

In [6] a pixel wise classification method based on dynamic bayesian network for vehicle detection is proposed. In [7] an object detection scheme is presented which identifies changes in image series. Foreground vehicle segmentation using gaussian mixture model is highlighted in [8]. An adaptive background model having frames averaging with respect to time is discussed in [9]. For vehicle localization ResNet model is used in [10]. A vehicle classification system involving deep learning is presented in [11]. An integrated vehicle detection and classification method is discussed in [12]. [13] discusses YOLOv3 algorithm for vehicle detection. [14] presents receptive field based neural network. Semantic image segmentation is used in [15]. Mask R-CNN with transfer learning is used in [16]. [17] discusses YOLO based solution. A review on vehicle detection is highlighted in [18]. In [19] deep learning assisted vehicle segmentation is discussed.

In [20] vehicles in airborne images are detected. An ensemble based method using image descriptors is discussed in [21]. In [22] and [23] methods are developed through application of gaussian mixture model. In [24] support vector machines is used for vehicle detection. SIFT algorithm is integrated with support vector machines in [25] to achieve vehicle detection. For autonomous vehicles an object detection system is highlighted in [26].

Some R-CNN version of vehicle detection algorithms are discussed in [28], [44] and [54]. Several significant YOLO based multi object vehicle detection algorithms are highlighted in [27], [32], [33], [34], [35], [37], [38], [39] and [42]. Vehicle detection algorithms in different weather conditions are presented in [29], [30] and [52]. Deep learning networks for vehicle detection are used in [31], [36], [41], [43], [45], [46], [55] and [56]. Several other notable works are available in [40], [47], [48], [49], [50], [51], [53] and [57].

## 3 Computational Method

In this section, smart traffic vehicle management using Faster R-CNN based deep learning based ensemble method is highlighted. The research problem revolves around traffic management which involves vehicle segmentation at different background levels [5]. In order to achieve this, we study Faster R-CNN [5] which analyses vehicles in smart traffic. The segmentation activity on various strategic areas of smart traffic analytics provide information related to decision-making activities.

### 3.1 Datasets used

The experimental datasets are adopted from [58]. The datasets are prepared keeping in view different traffic conditions. At the very first instance multi-class vehicle objects are considered. Several challenge factors such as traffic jams and overlapping vehicles are incorporated in dataset. Broadly speaking datasets are developed with respect to two different situations viz high density and low density traffic scenes. The former comes with several objects in single image while later has single class per image having no overlapping. In order to achieve better training, images from both above mentioned situations are placed in different datasets. The high density traffic scenes are considered from several places having traffic which are less crowded in nature. A total of 1000 images from six classes of vehicles viz two wheelers, three wheelers, four wheelers, six wheelers, eight wheelers and ten wheelers are developed. Figures 2(a) and 2(b) highlights some sample images. High density traffic scenes are considered from daily traffic which are congested in nature. 5000 images from above stated classes are created. Certain critical factors such as illumination, occlusions etc are incorporated irrespective of appearance, shape, scale and size in this dataset. Table 1 highlights certain high and low density dataset statistics alongwith annotations of class. These datasets are preprocessed and augmented as discussed in subsection 3.2. Each of these datasets are benchmarked with respect to standard datasets [58].

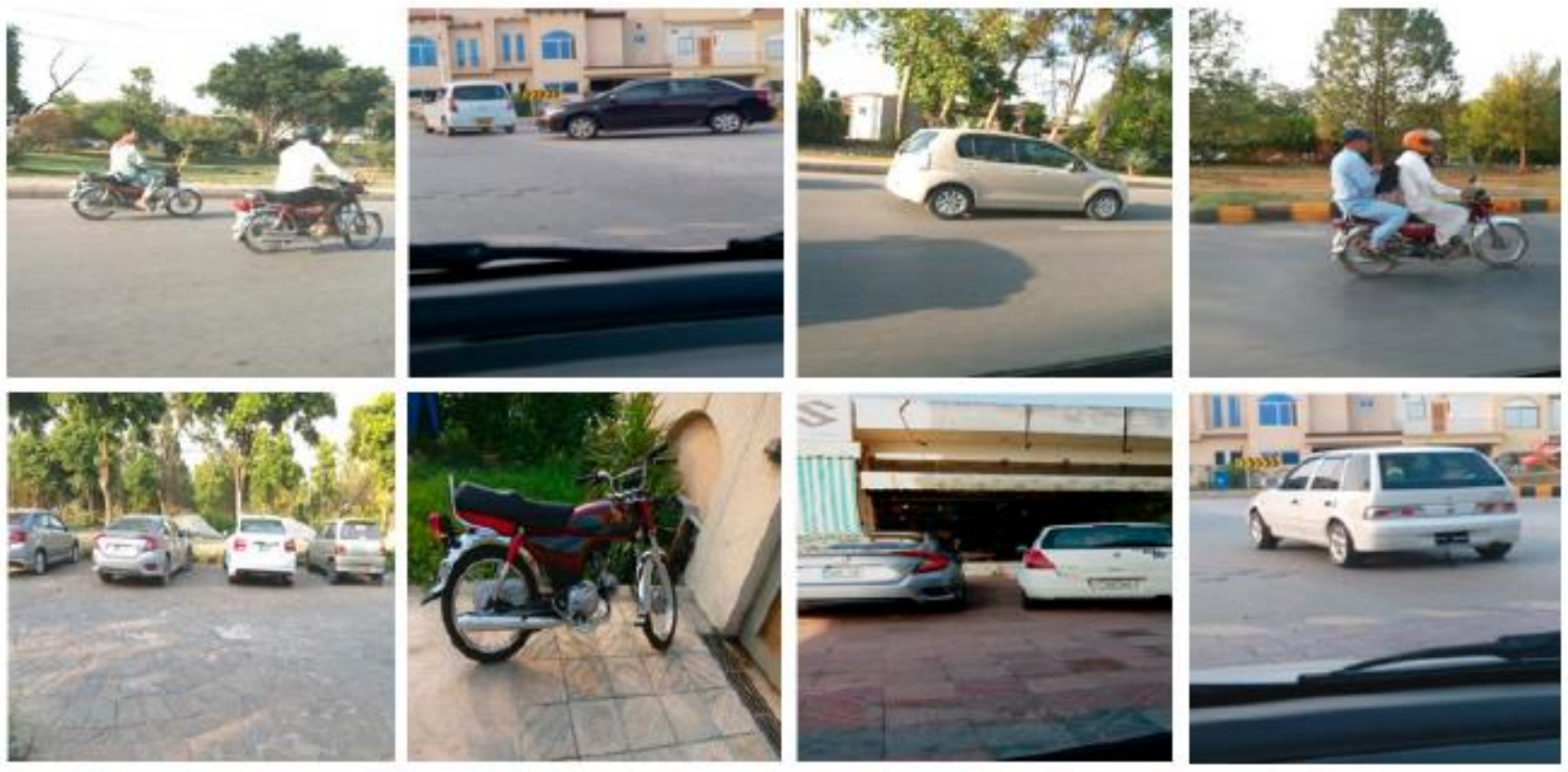

Figure 2(a): Sample images from dataset - low density traffic

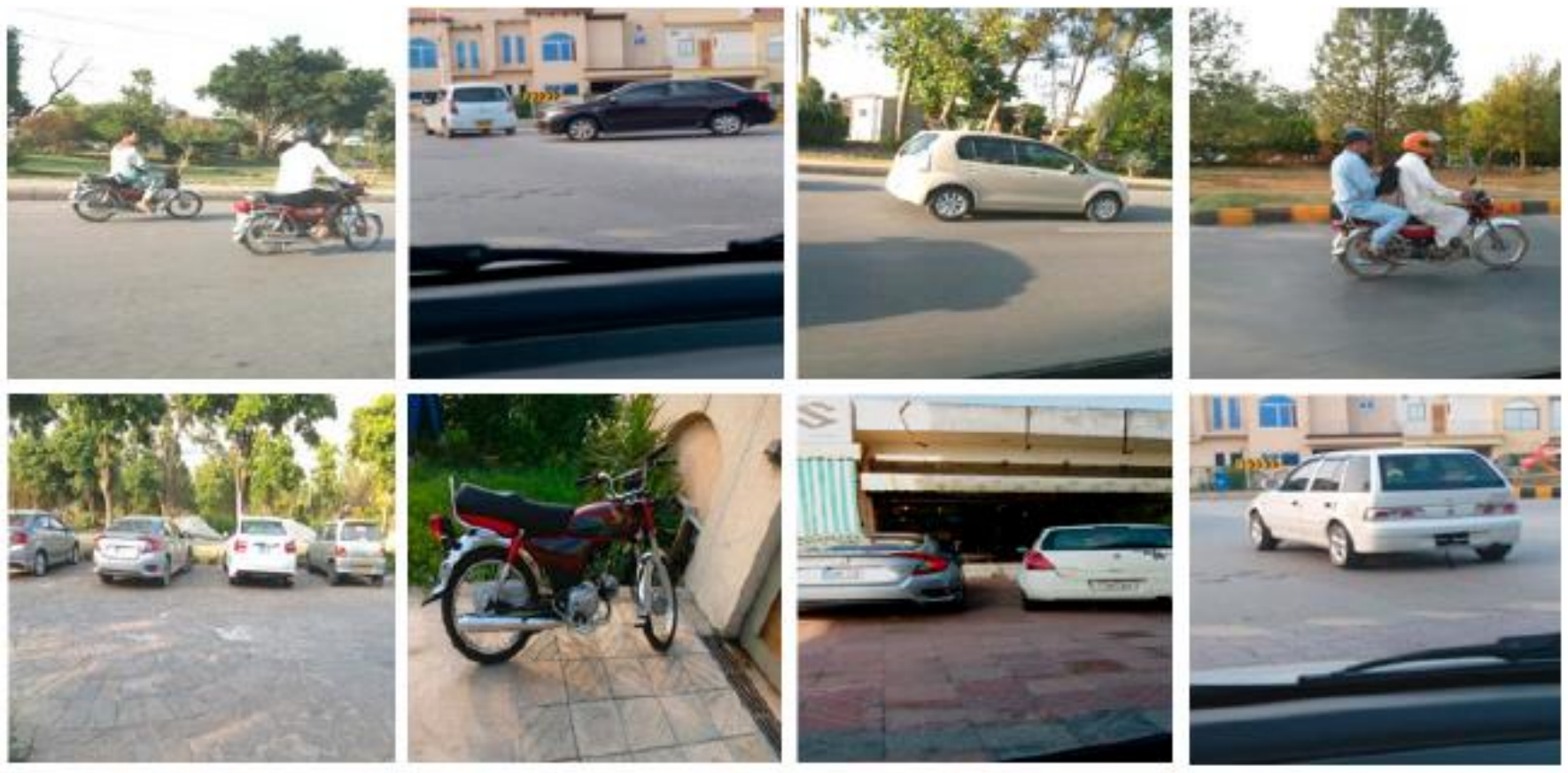

Figure 2(b): Sample images from dataset - high density traffic

| | | | | |
|---|---|---|---|---|
| | Dataset | | low density traffic | high density traffic |
| | Source Images | | 5000 | 1400 |
| | Annotations | | 50595 | 2000 |
| Classes | | two wheeler | 9000 | 120 |
| | | three wheeler | 9351 | 200 |
| | | four wheeler | 10879 | 880 |
| | | six wheeler | 10500 | 600 |
| | | eight wheeler | 5990 | 110 |
| | | ten wheeler | 4875 | 90 |

Table 1: Statistics of dataset images

### 3.2 Data Annotation and Augmentation

In order to achieve reliable vehicle detection dataset classes are labelled. Based on motivation from [5], an image tool [57] has been used here towards labeling and annotation of this dataset. As shown in Figures 3(a) and 3(b) for each object in image, a bounding box is assigned manually. In high density datasets (HDD) from high density traffic many bounding boxes are present. In low density datasets (LDD) from low density traffic few bounding boxes are present in single image. The label of respective class is specified by these bounding boxes. As mentioned in subsection 3.1, the whole dataset is defined through six classes.

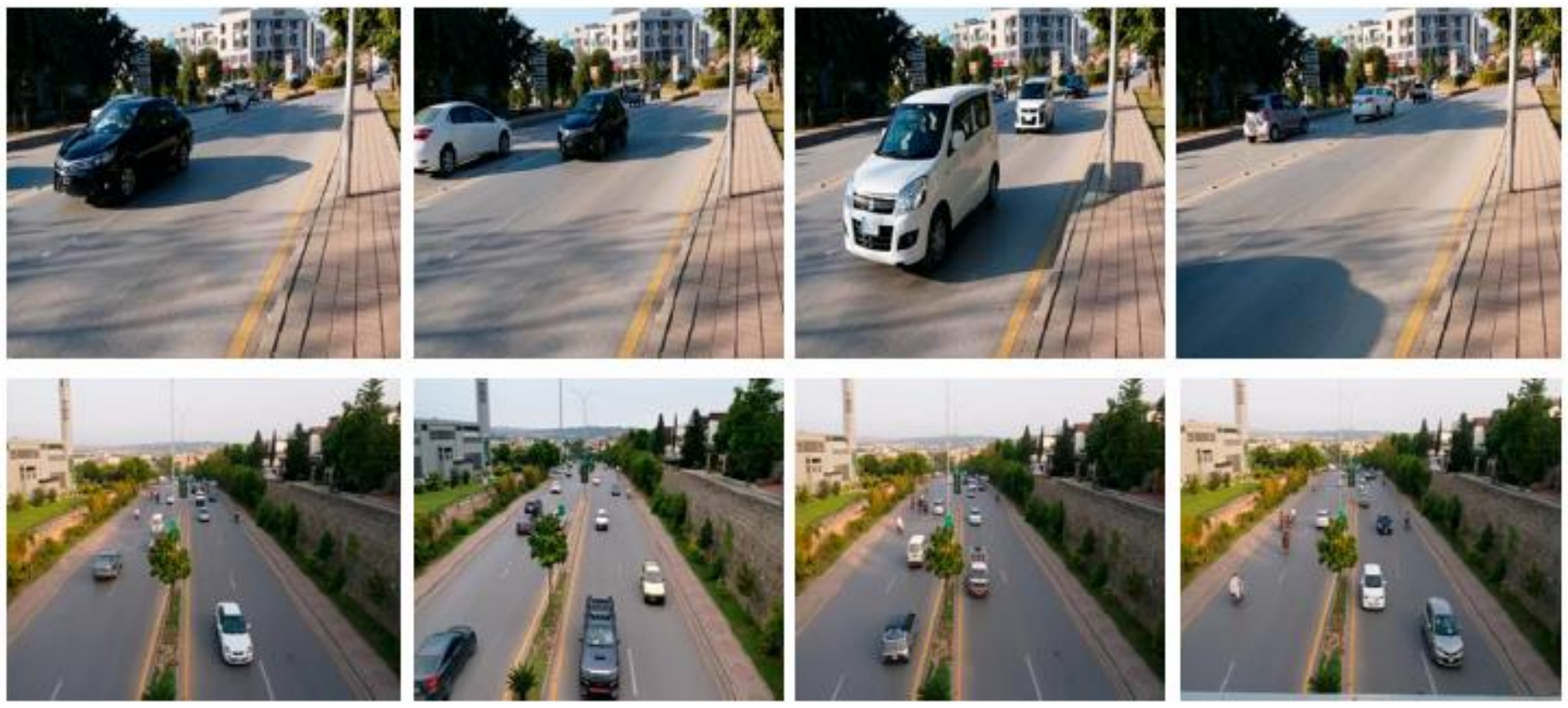

Figure 3(a): Sample annotated images from dataset - video dataset

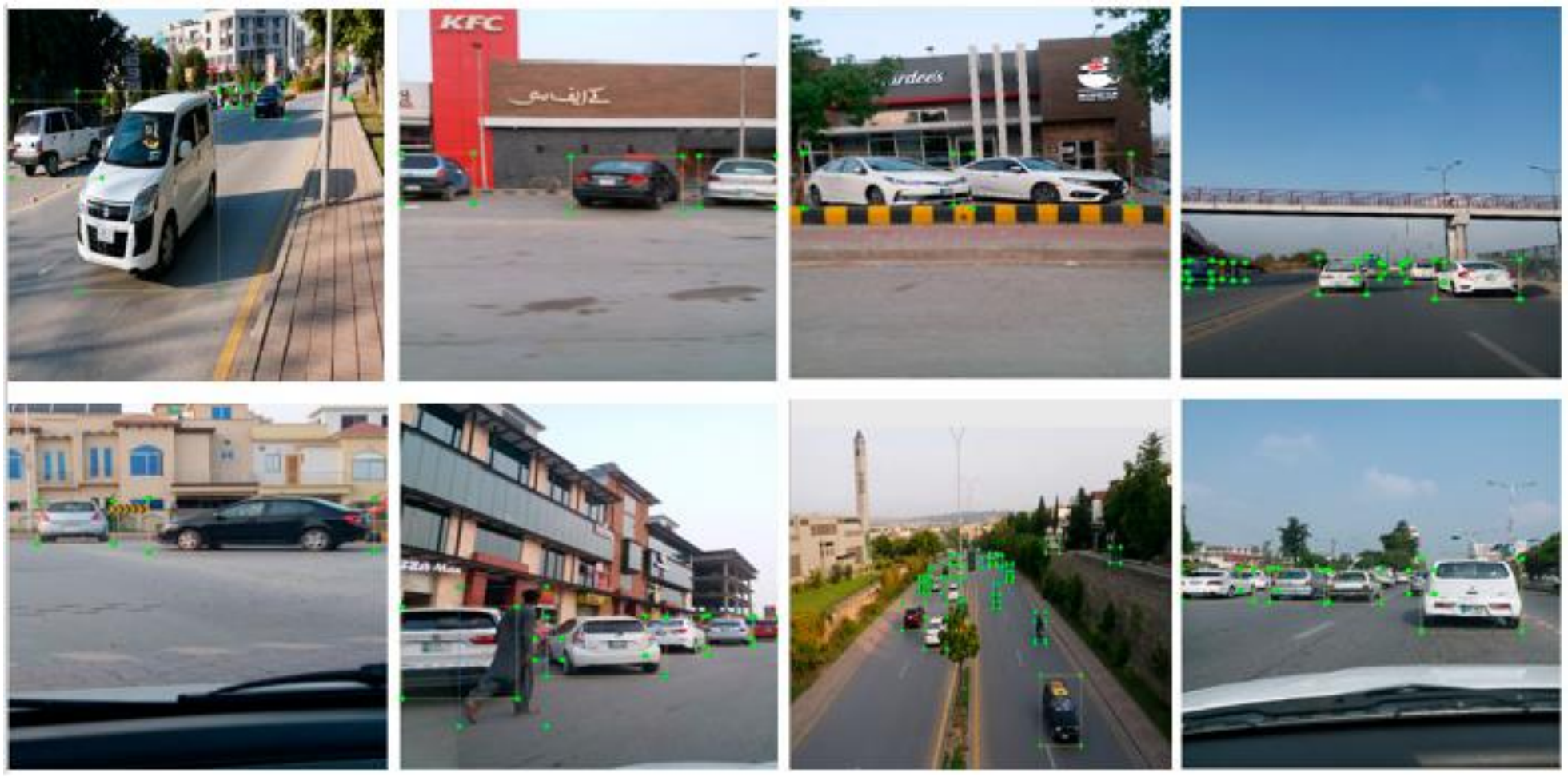

Figure 3(b): Sample annotated images from dataset - annotated images

In order to increase features of datasets such that better results are obtained data preprocessing and augmentation is used. These addresses various issues which can be present in images such as noise, inconsistency and unbalanced classes. The data augmentation process used here are in lines with discussed in [5] and [57]. Since augmented dataset has an unknown distribution it is benchmarked adhering to certain standards [58].

### 3.3 Faster R-CNN based Method

Now we present a detailed description of proposed method. The method is highlighted considering four steps [5] viz (a) minimization with adaptive background model (b) Faster R-CNN based subnet operation (c) Faster R-CNN initial refinement and (d) result optimization with extended topological active nets.

Adaptive background modeling is used to construct background with traffic video as input. The video frames are analyzed in order to develop background model. The objective here is to find best background estimate. With this on foreground model shadow and illumination change impacts are minimized. This process is initiated by first initializing few frames and then continuous updates are done each time. This leads to extraction of foreground from next set of frames. The initialization of background model is done considering first frame's pixel values. A subsequent update of model is performed through calculation of value of pixel considering background model:

$$pixel_j(y) = \overline{pixel}_{j-1}(y) + \frac{1}{GP_j\sqrt{2\pi\sigma^2}} \exp\left(\frac{-1}{2}\left(\frac{y-y_j}{\sigma}\right)^2\right) \quad (1)$$

Here $y_j$ represents value of pixel with respect to $j^{th}$ frame. The value of $\overline{pixel}_j$ is computed as follows:

$$\overline{pixel}_{j-1}(y) = \frac{pixel_j(y)}{\sum_j pixel_j(y)} \quad (2)$$

Here number of frames considered is $N$. The gain parameter is denoted by $GP_j$. It takes care of background modeling's learning rate. The new information is learned by increasing $GP_j$ as presented in Equation (3) with prior information disappearing slowly.

$$GP_j = Gain\left(\frac{2}{1+\exp\left(-\frac{cont-\alpha}{\beta}\right)}\right) \quad (3)$$

Here $Gain$ and $\alpha$ parameters take care of sinusoid function's inflection point. The parameter $\beta$ controls gradient. The parameter $cont$ depends on number of frames. Background model is updated for every frame. Considering every fame with background as adaptive, after its subtraction we reach to foreground objects.

After adaptive background model based minimization is performed, we present basic Faster R-CNN architecture used in this research. The architecture of Faster R-CNN is considered from [5] having variation based baseline considered from [70]. The architecture is highlighted in Figure 4. In order to validate proposed method, datasets mentioned in subsection 3.1 are used.

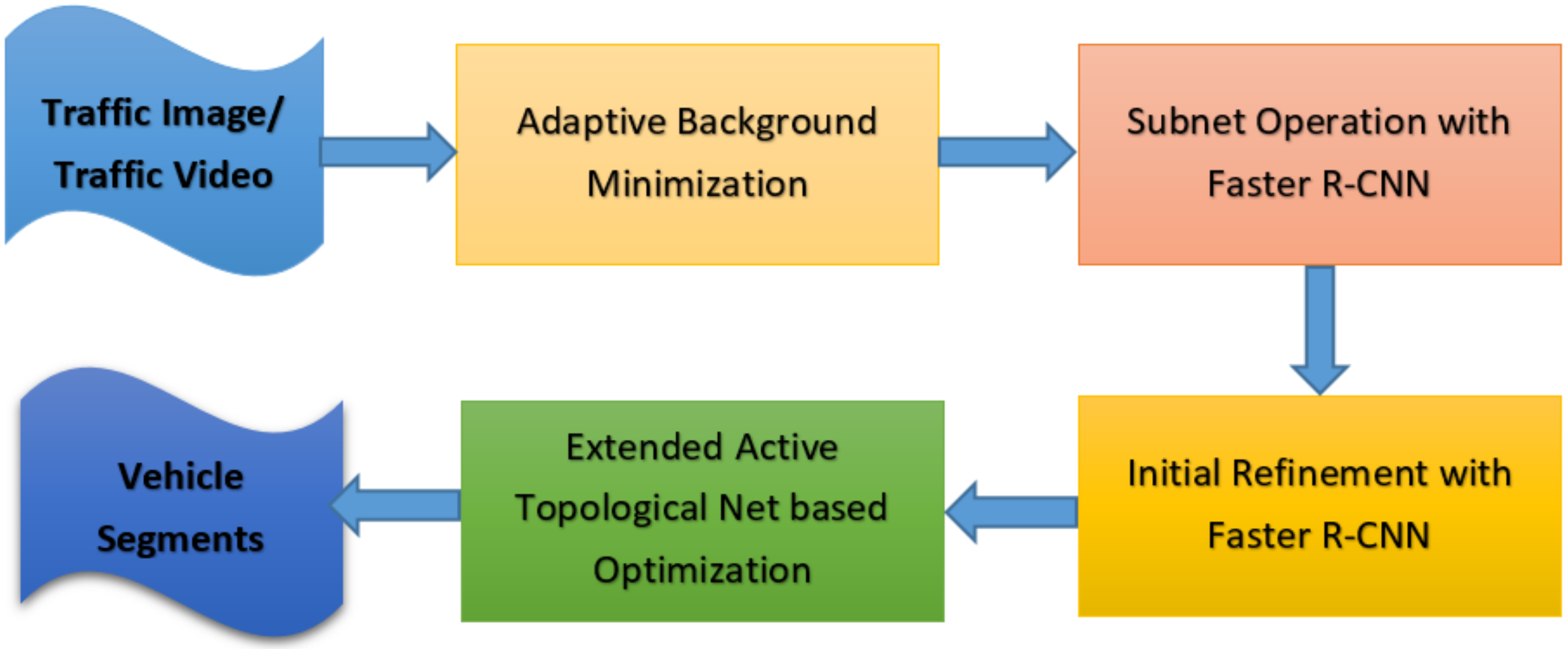

Figure 4: Architecture of vehicle segmentation method

The convolutional feature map is developed when entire image is processed convolutional and max pooling layers. For each object's RoI pooling layer, feature vector of fixed length is extracted. The input of sequence of fully connected layers are feature vectors. From a set of fully connected layers, output passes to sibling layers where softmax probability estimates are produced. The object classes estimate softmax probability. Here background class with set of other layers are considered. The encoding is performed on set of values with respect to refined positions in bounding box for each object class.

For RoI pooling layer we use max pooling. This allows conversion of features into small feature map. Here convolutional feature map based RoI is defined through 4-tuple. RoI max pooling divides rectangular window into sub window grids. The channel independent pooling is applied for each feature map. The network weights are trained through backpropagation in Faster R-CNN. A hierarchical sampling of RoIs is performed for each image. Stochastic gradient descent (SGD) training for Faster R-CNN is done in mini-batches. For training with larger datasets, execution of SGD is done for more iterations.

Faster R-CNN makes use of sibling output layers. For each RoI we have discrete probability distribution as initial output. The probabilities for fully connected layer have outputs as softmax. The labeling for RoI is performed with respect to ground truth class. For each training bounding box regression having ground truth is considered. For each labeled RoI considering multitask loss, joint classification is present with respect to training and bounding-box regression. The optimization of multitask loss is performed as highlighted in [5]. For whole image classification, convolutional layers' calculation time is greater than fully connected layers. RoIs processing time is appreciably large for detection.

Now subnet operation with Faster R-CNN is presented. The foreground image considers a mesh. This is moved towards subnet operation for Faster R-CNN. The training image dataset is adopted from [19]. This is marked with respect to active net grids. The subnet is represented is binary matrix with 1 as indicator for link presence of mesh and 0 as indicator for link absence. Here Faster R-CNN is trained where features are learned in order to produce ground truth specific results. The U-Net based CNN [59] helps us to perform initial subnet operation.

The extended topological active net refines initial subnet. Faster R-CNN produces hole based mesh considering all background. The objects have mesh nodes. Those mesh nodes which where boundary is not crossed are removed. It is reached though extended topological active net energy minimization. The energy function used here is highlighted in Equation (4). By using Equation (1) adaptive background model is developed.

$$Energy\big(w(a,b)\big) = \int_0^1 \int_0^1 \big[Energy_{internal}\big(w(a,b)\big) + Energy_{external}\big(w(a,b)\big)\big]\, dadb \quad (4)$$

The mesh deformation is done using greedy approach. In situations with complex deformation of shapes, extended topological active net produces different energy term. This it achieves through various thresholds. When clutters and occlusion are present in vehicle image segmentation extended topological active net always reaches local minima. This results in low segmentation accuracy. The improved version of extended topological active net provides resolution to these problems. The combined effect of all models used here leads us to better results.

### 3.4 Training Process

The training process is now briefly discussed. The annotated and augmented data is trained using Faster R-CNN algorithm. In order to perform training, several parameters like size of batch, epochs needed and resolution of image are used. Since network is trained from scratch, random weight initialization is performed. Here initially trained COCO weights [5] are used towards model training with appreciable time and computation benefits. The best weight values are obtained using initially trained Faster R-CNN having transfer learning. [5] datasets are used as benchmark in order to train stated custom datasets. The batch sizes of 5, 10, 20, 30 and 40 are considered. The epochs are also changed to 100, 200, 300, 400 and 500. The confidence values are considered between 0.4 to 0.6. The best weights are used to detect objects in datasets. The predicted labels and assessment images incorporating bounding boxes with confidence values are also obtained.

### 3.5 Evaluation Criteria

In this research precision and recall criteria [5] are used for measuring robustness. Here, mean average precision (mAP) is also calculated for recall values lying between 0 to 1.

## 4 Experiments and Results

Here a detailed discussion on simulation results is presented. We conducted detailed experiments in Google Colab having T4 GPU with Intel Xeon CPU and 64 GB of RAM. Python version 3.11.5 has been used as simulation tool in this research. In order to assess vehicle detection method performance, several state-of-the-art detectors are evaluated. Also various comparisons are performed with stated method considering accuracy and execution times. All methods are trained on data adapted from COCO [60] and DAWN [61] datasets.

| Dataset | Number of Classes | Training | Validation | Testing |
|---|---|---|---|---|
| HDD | 6 | 16799 | 20575 | 20685 |
| LDD | 6 | 80577 | 10879 | 10780 |
| COCO | 90 | 300000 | 18000 | 12000 |

Table 2: Certain dataset specific statistics
(HDD: High density dataset from high density traffic;
LDD: Low density dataset from low density traffic)

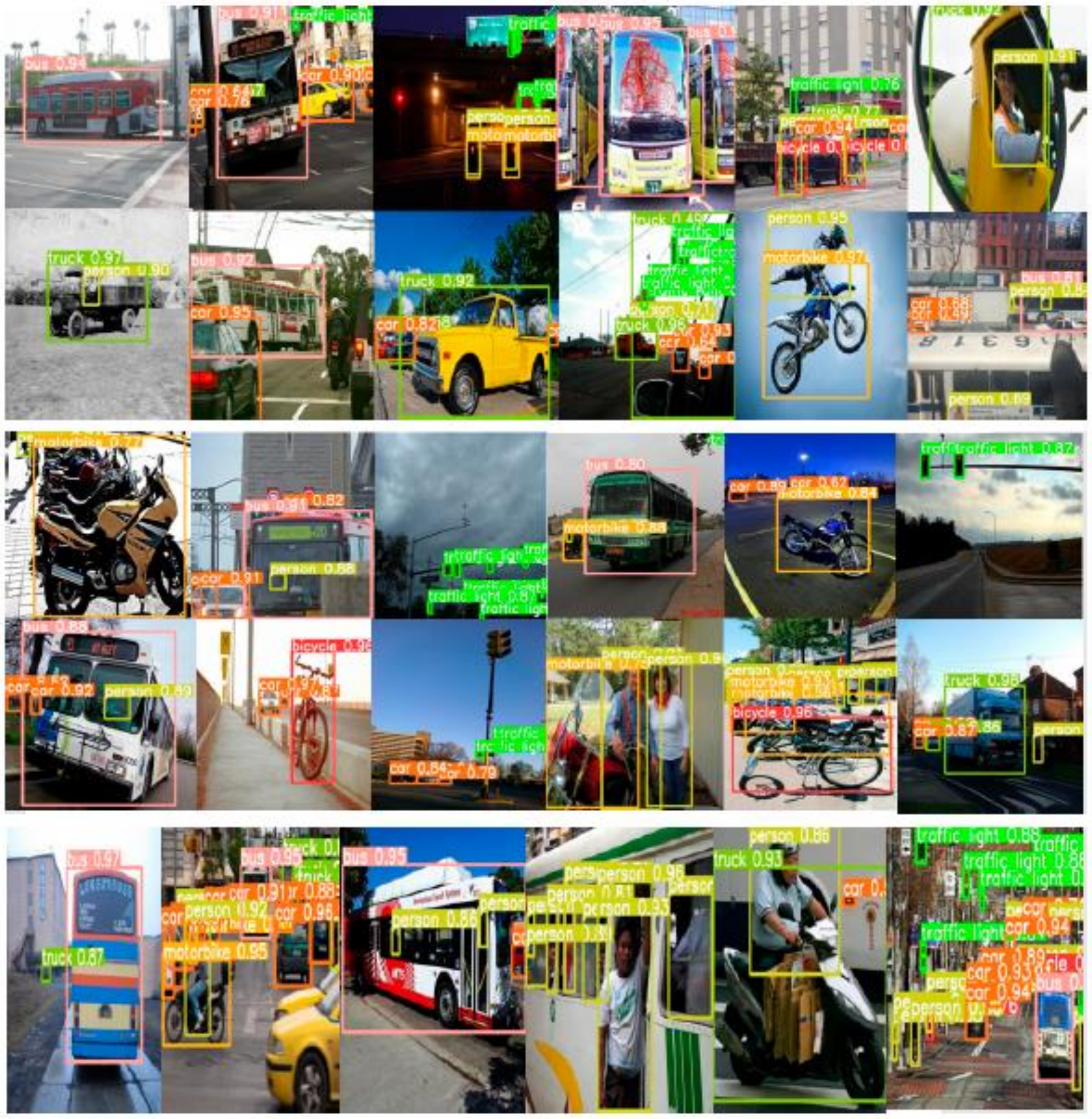

Figure 5: COCO dataset vehicle detection results

| Computation Backbone | Input Size | Multi-Scale | mAP (%) |
|---|---|---|---|
| CSPDarkNet53 [42] | 512 × 512 | False | 48.64 |
| CNN [43] | 512 × 512 | False | 49.00 |
| R-CNN [44] | 512 × 512 | False | 47.40 |
| BottlenectCSP [45] | 512 × 512 | False | 28.99 |
| VGGNet-16 [47] | 512 × 512 | False | 30.40 |
| ResNet-101-FPN [48] | 512 × 512 | False | 40.40 |
| VGGNet-16 [49] | 800 × 800 | False | 42.00 |
| ResNet-101 [50] | 800 × 800 | False | 49.40 |
| ResNet-101 [51] | 512 × 512 | False | 40.50 |
| CNN + SVM [52] | 512 × 512 | False | 50.10 |
| BN + ReLU [53] | 512 × 512 | False | 34.90 |
| ResNet-C4-FPN [54] | 512 × 512 | False | 32.88 |
| ResNet-50 [55] | 512 × 512 | False | 50.90 |
| SiNet [56] | 512 × 512 | False | 51.50 |
| CSP [57] | 512 × 512 | False | 52.45 |
| Our Method | 512 × 512 | False | 58.90 |

Table 3: COCO dataset accuracy comparison

The detection and segmentation of objects is performed by COCO dataset considering natural contexts [5]. In Table 2 COCO dataset highlights several objects collected from complex scenes. The dataset has images of 100 different object types with 3 million instance labels. The results are highlighted in Figure 5. Certain YOLOv5 semantics are adopted from [57]. All vehicles are accurately detected by stated method considering variation in illumination. In order to further validate superiority of stated method we perform a comparative analysis with 14 methods as presented in Table 3. Some significant observations are briefly highlighted here.

For images with different resolutions, stated computational structure provides best performance in terms of mAP values. The methods presented in [49], [50], [55], [56] also provide promising results with respect to COCO datasets. [54] makes use of ResNet and produces results on lower side for object detection in COCO datasets. Similar results are obtained from methods presented in [45] and [47]. Here miscellaneous objects are detected on COCO datasets with promising results. These include vehicles of varying shapes and sizes. All results are highlighted in Figure 5.

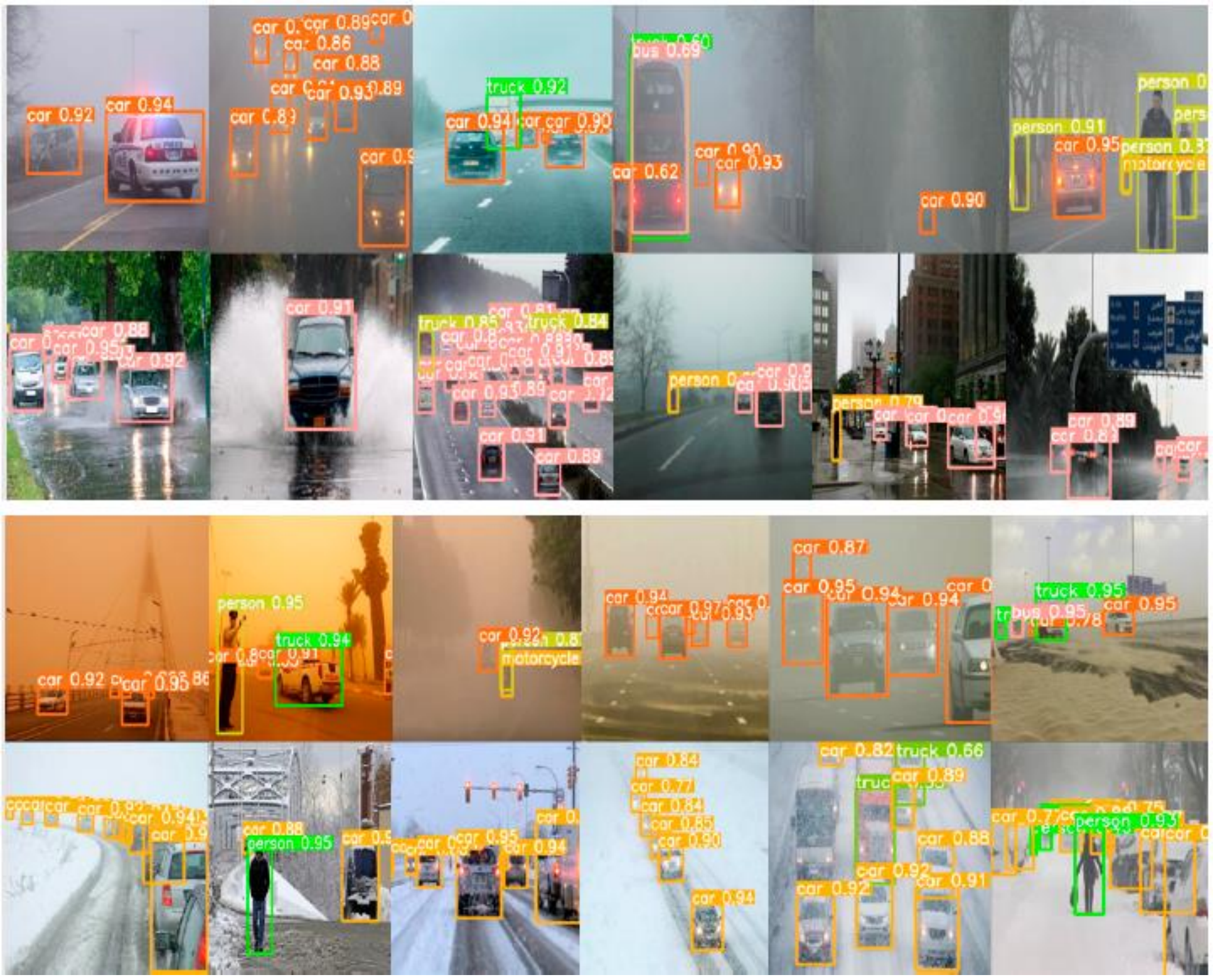

Figure 6: Vehicle detection results on DAWN datasets

The DAWN dataset is used in order to study and investigate performance of stated method. DAWN dataset has 2000 image with different variations [5]. It shows varying traffic situations in different weather conditions. Figure 6 shows fairly detailed results considering significant observations. The stated method addresses all prevailing weather situations such as rainy days, normal dry days and snowy days.

A comparison is performed with results presented in Tables 4. Some observations are highlighted here. In fog situation [54] has highest rank. Our method exceeds results of [54]. Here [43] and [52] have lowest ranks. In rain situation our method produces best results followed by [57], [45], [48], [50] and [51]. Here [52] has lowest rank. In snow situation our method has highest rank followed by [51] and [57]. Here [45], [47], [50], [51] produce similar results. In dry day situation stated method achieves highest rank. Here [42], [48], [49] yield similar results. [52] has lowest rank.

Now we present some additional insights in this research with respect to diverse range of environments. All state-of-the-art object detection methods have been studied in this research. These results are briefly discussed here.

| Computation Backbone | Fog | Rain | Snow | Dry Day |
|---|---|---|---|---|
| CSPDarkNet53 [42] | 26.40 | 31.55 | 39.95 | 24.10 |
| CNN [43] | 24.00 | 21.10 | 38.32 | 23.80 |
| R-CNN [44] | 27.20 | 21.30 | 28.30 | 18.00 |
| BottlenectCSP [45] | 29.31 | 41.21 | 43.00 | 24.02 |
| VGGNet-16 [47] | 23.40 | 24.60 | 37.90 | 15.83 |
| ResNet-101-FPN [48] | 28.95 | 41.10 | 43.00 | 24.09 |
| VGGNet-16 [49] | 23.10 | 27.65 | 34.00 | 24.10 |
| ResNet-101 [50] | 29.70 | 40.10 | 43.00 | 23.99 |
| ResNet-101 [51] | 28.10 | 40.40 | 43.02 | 24.10 |
| CNN + SVM [52] | 16.50 | 14.08 | 15.38 | 10.69 |
| BN + ReLU [53] | 25.08 | 19.14 | 23.18 | 17.38 |
| ResNet-C4-FPN [54] | 29.68 | 30.32 | 33.93 | 24.00 |
| ResNet-50 [55] | 28.83 | 27.68 | 30.19 | 24.03 |
| SiNet [56] | 26.45 | 20.09 | 27.92 | 11.31 |
| CSP [57] | 29.66 | 41.21 | 43.01 | 24.14 |
| Our Method | 30.40 | 45.55 | 45.79 | 25.60 |

Table 4: Comparison of methods on DAWN datasets

In [42] BIT-Vehicle and UADETRAC datasets are investigated. In [57] 3 different datasets are studied which several variations with respect to road conditions, weather as well as complex background. In [43] and [48] primary concentration is on KITTI and DAWN datasets with certain deviations which produces appreciable results. In [44] investigation of object detection capability is performed on COCO and DAWN datasets. In [45] CARLA dataset is explored. The results presented here extends method's detection capability to 3 other datasets. PASCAL VOC 2007 dataset is studied in [47] and [51]. In [37] PASCAL dataset is studied which contains annotated images of various objects. Here detection capability of method is expanded to 3 different vehicle datasets. In [49], [50], [53] and [54] COCO and DAWN datasets are used in order to validate methods for object detection. In [54] various object detection methods are presented for detection of different objects. PETS2009 and changedetection.net 2012 datasets are studied in [52] with DLR Munich vehicle and VEDAI datasets in [55] and KITTI and LSVH datasets in [56]. The vehicle detection methods on these datasets produce appreciable results.

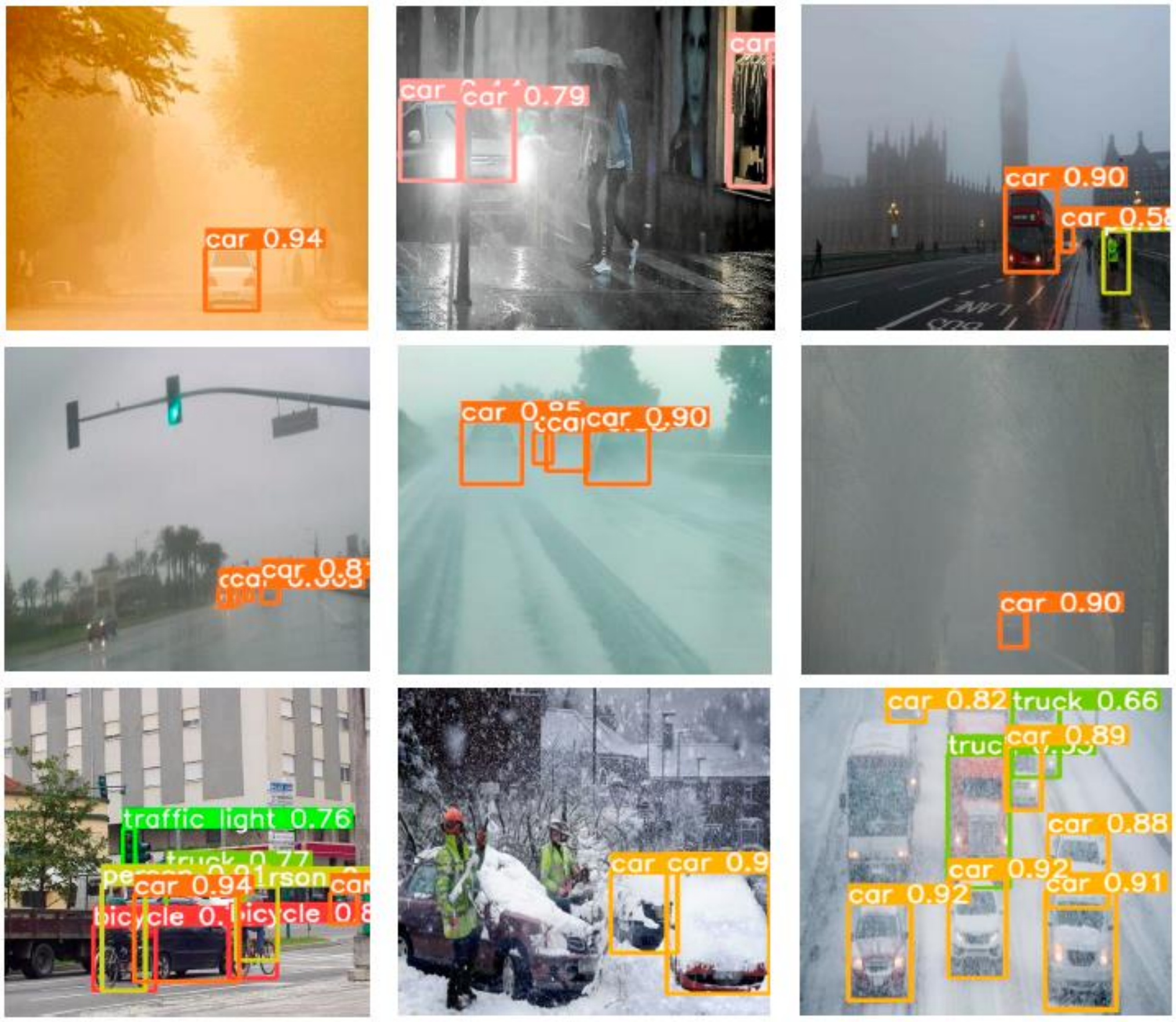

Figure 7: Vehicle detection method on sample images

In DAWN dataset has more challenging images with fog, rain, dry day and snow. However, as shown in Figure 7 very low vehicle detection results are achieved. But our method performs well in this situation. Considering several state-of-the-art approaches, a detailed comparison of is performed alongwith our method. It is seen that in different situations our method performs well. In COCO dataset our method produces appreciable results in comparison to other methods.

## 5 Conclusion

In this research work smart traffic management of vehicles is studied using Faster R-CNN based deep learning method. It is an intractable problem in computer vision and artificial intelligence domain. When occlusions, background with clutters and traffic with density variations are present, this problem becomes more challenging. The computational paradigm involves four steps viz minimization with adaptive background model, Faster R-CNN based subnet operation, Faster R-CNN initial refinement and result optimization with extended topological active nets. The concept of adaptive background modeling is incorporated in this framework. The shadow and illumination related issues are also addressed. The topological active net deformable models help to achieve higher segmentation accuracy. The deformations are reached with topological and extended topological active nets. Mesh deformation helps in minimization of energy. The segmentation accuracy is improved with modified version of extended topological active net. The superiority of this method is highlighted with experimental results.